\documentclass[conference]{IEEEtran}
\IEEEoverridecommandlockouts
\usepackage{cite}
\usepackage{amsmath,amssymb,amsfonts}
\usepackage{algorithmic}
\usepackage{graphicx}
\usepackage{textcomp}
\usepackage{xcolor}
\usepackage{hyperref}
\usepackage{graphics}

\def\BibTeX{{\rm B\kern-.05em{\sc i\kern-.025em b}\kern-.08em
    T\kern-.1667em\lower.7ex\hbox{E}\kern-.125emX}}
\begin{document}

\newcommand{\ron}[1]{{\textcolor{blue}{[Ron: #1]}}}
\newcommand{\asaf}[1]{{\textcolor{red}{[Asaf: #1]}}}
\newcommand{\guy}[1]{{\textcolor{orange}{[Guy: #1]}}}
\newcommand{\moshe}[1]{{\textcolor{gray}{[Moshe: #1]}}}

\title{GIM: Gaussian Isolation Machines\\
}

\author{\IEEEauthorblockN{Guy Amit, Ishai Rosenberg, Moshe Levy, Ron Bitton, Asaf Shabtai, and Yuval Elovici}
    \IEEEauthorblockA{\textit{Department of Software and Information Systems Engineering} \\
    \textit{Ben-Gurion University of the Negev}\\
    Beer-Sheva, Israel \\
    \{guy5,ishairos,moshe5,ronbit\}@post.bgu.ac.il, \{shabtaia,elovici\}@bgu.ac.il }
}

\maketitle

\begin{abstract}
In many cases, neural network classifiers are likely to be exposed to input data that is outside of their training distribution data.
Samples from outside the distribution may be classified as an existing class with high probability by softmax-based classifiers; such incorrect classifications affect the performance of the classifiers and the applications/systems that depend on them.
Previous research aimed at distinguishing training distribution data from out-of-distribution data (OOD) has proposed detectors that are external to the classification method.
We present Gaussian isolation machine (GIM), a novel hybrid (generative-discriminative) classifier aimed at solving the problem arising when OOD data is encountered.
The GIM is based on a neural network and utilizes a new loss function that imposes a distribution on each of the trained classes in the neural network's output space, which can be approximated by a Gaussian.
The proposed GIM's novelty lies in its discriminative performance and generative capabilities, a combination of characteristics not usually seen in a single classifier. 
The GIM achieves state-of-the-art classification results on image recognition and sentiment analysis benchmarking datasets and can also deal with OOD inputs.
% We also demonstrate the benefits of incorporating part of the GIM's loss function into standard neural networks as a regularization method.
\end{abstract}


\section{Introduction}
\label{section:one}
\begin{figure}[t]
\centerline{\includegraphics[scale=0.47]{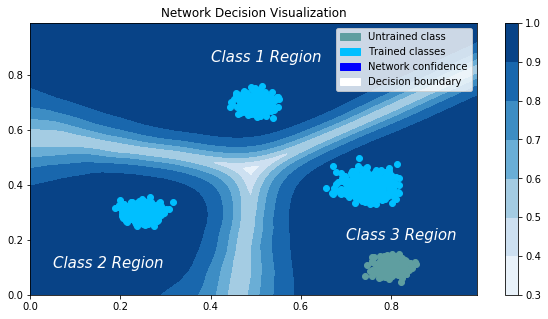}}`
\caption{Neural network input space trained on three classes (in light blue) predicting a fourth class (in green) in the class 3 region.}
\label{figure1}
\end{figure}
\par
In recent years, neural networks have successfully been used for classification tasks in various domains for numerous tasks, including computer vision~\cite{he2016deep, xu2015show, lawrence1997face}, natural language processing~\cite{chopra-etal-2016-abstractive, lample2016neural,dos-santos-gatti-2014-deep}, voice recognition~\cite{venayagamoorthy1998voice}, and even in the domain of information security for malware detection (classifying files as malicious or benign)~\cite{rosenberg2018generic}. 

Neural networks are known for their ability to learn complex patterns. 
As a result, they are better at representing complex domains than standard machine learning algorithms.
The use of domain-specific operators, e.g., pooling layers, convolution layers, and recurrent neural network (RNN)  cells~\cite{cnn2019survey, bengio1994learning, Pearlmutter1989LearningSS}, has allowed neural networks to fit the data easily and serve as accurate classifiers. 

The use of softmax classification layers (a dense layer paired with a softmax activation function)~\cite{dunne1997pairing,nwankpa2018activation} is a common practice in neural network classifiers. 
The softmax classification layer is a linear classifier, with respect to the previous layers of the neural network.
Softmax layers are used due to their probabilistic outputs. 
According to~\cite{dunne1997pairing}, pairing the softmax layer with the cross-entropy loss provides improvements in convergence speed which aren't seen when the output layer is paired with other types of loss functions.
The softmax layer uses straight lines to draw the decision boundary between the classes, whereas all of the other layers extract features. 
The decision boundary drawn by the softmax layer divides the space into areas, such that each class has its own area.
Given a new sample, the network tries to determine which area the sample belongs to, thereby classifying the sample accordingly.
This study is aimed at better understanding the extent to which the area assigned to a class is ``actually the class itself.''

Figure~\ref{figure1} illustrates the input space of a neural network trained on data sampled from three two-dimensional Gaussians. 
The colors in the figure indicate the network's confidence in the prediction; the darker the color, the more confident the network is regarding the predicted class. 
The instances used for training the network appear in each class region. The following three observations can be made based on the figure. 
First, large regions in the decision space do not contain input data from any of the three classes, however the probability output of these areas is high (dark blue).
Second, when the network is presented with a fourth class (the group of points in green, termed the ``untrained class'') which are samples from unknown distribution, the network classifies it with high confidence as one of the trained classes, instead of issuing an alert stating that it has encountered data from an unknown distribution. 
Third, no area between classes represents instances of other classes, even the decision boundary has a probability of 0.3 of belonging to one of these classes.
These three observations demonstrate that neural networks do not actually capture the distribution of each class in the training set but rather learn how to differentiate between classes without considering the possibility that the input instance may (mistakenly or even intentionally) not belong to the classes in the training set. 
The main problem with this type of learning, i.e., softmax-based learning, is that the classifier does not consider the possibility that other types of data might exist. 
As in Figure~\ref{figure1}, the fourth class is recognized as class 3 with high confidence.
In real-world scenarios such phenomena frequently occur (e.g., when an unexpected object appears in front of an autonomous vehicle or a facial recognition system tries to recognize a person with his/her head tilted, causing the classification probability to be indecisive, i.e., for different people to have a similar probability).
%something is weird with here, and if I create a new paragraph the whole test shifts
There are some methods that aimed to solve this problem. However, most of them are doing so by adding external components to the classifier. Our goal was to create a classifier that has the capability of detecting OOD data intrinsically and would have performance comparable to those of discriminative models.

In this paper, we propose a hybrid classification method which is based on neural networks and a new loss function that aims to solve the mentioned problem caused by the softmax layer.
Our approach utilizes concepts of generative and discriminative modeling~\cite{ng2002discriminative} to create a hybrid classification method with a built-in confidence metric that enables it to deal with data from other distributions.

We evaluate our method on four datasets (three computer vision benchmarking datasets and one sentiment analysis datsaset) and various neural network architectures, and show that it can achieve accuracy comparable to that of standard neural networks and is capable of dealing with data outside of the trained distribution, without employing additional anomaly detection algorithms or input prepossessing.

The main contributions of this paper is:
a neural network-based hybrid classifier with state-of-the-art accuracy. The classifier's accuracy is similar to that of discriminative models, while being inherently capable of identifying data from other distributions, and like generative models, the proposed classifier calculates a confidence score for its predictions. 
% (2) We present two new representation based regularization terms that improve the classification performance of standard neural networks.

\section{Related work}
\label{section:two}
\subsection{Generative Classifiers vs Discriminative Classifiers}
Machine learning classifiers are often divided into two families: generative and discriminative.
The difference between the two is the information produced when calculating the prediction.

Formally, let {$x$} be a sample and {$y$} be a label. 
Generative classifiers learn a model of the joint probability $p(x,y)$ and make a prediction by calculating $p(y|x)$.
This enables {$p(x|y)$} to be calculated as well.
Discriminative classifiers do not calculate {$p(x|y)$}; instead they predict the posterior probability {$p(y|x)$} directly~\cite{ng2002discriminative}.  
The term {$p(x|y)$} can be interpreted as a confidence rate for the prediction, i.e., the probability for ${x}$ from the input space to be labeled as a specific ${y}$, which is used as a measure of 
${y}$ being the correct prediction for ${x}$. 
Although the confidence rate can be useful in various applications, in practice, generative classifiers are not usually used due to the fact that they are outperformed by discriminative classifiers. 

\subsection{Identification of Out-of-Distribution Data with Neural Networks}

A classifier that can identify whether a sample is not from the same distribution as the training data is capable of handling unpredictable inputs. 
The presence of unpredictable inputs can be intentional or accidental.
Technically, identifying out-of-distribution (OOD) data means that the model labels the input as OOD instead of classifying it to a specific class. 
Classifiers based exclusively on softmax do not inherently have this capability, as softmax classifies every input to some class. 
Some research has been performed on unsupervised means of OOD detection, such as~\cite{chalapathy2018anomaly,an2015variational,chalapathy2017robust}. However, because the proposed methods use components that are external to the classifier, they require the training of an additional component/model for each class. 
Hendrycks \text{et al.}~\cite{hendrycks2016baseline} established a baseline method which is based on softmax. 
Later, Liang \textit{et al.}~\cite{liang2017enhancing} introduced ODIN. 
ODIN uses a distillation~\cite{hinton2015distilling} like softmax, combining it with adversarial-like perturbations~\cite{goodfellow2014explaining} to the input in order to predict whether the input is in or out of the distribution. 
This method does not require additional training, but it does require the performance of two feedforward and one backpropagation operations which makes it impractical for real-time use. 
The most recent research on OOD detection was conducted by Devries \text{et al.}~\cite{devries2018learning}. 
Their method adds an external functionality to the existing neural network design. 
In addition, their method is based on a neural network with softmax output, with the addition of an output neuron that serves as an OOD detector; the neuron added must be trained using a novel method which is described in the paper. 
In contrast, we introduce a method inherently capable of detecting OOD inputs.
We compare our results with the baseline proposed by Hendrycks \text{et al.}~\cite{hendrycks2016baseline}, which is the most closely related study performed on the subject, as it deduces the affiliation of the input to OOD, relying solely on representations learned by the classifier. 
\subsection{Multivariate Gaussians in Neural Networks}
Gaussians and neural networks are known for their ability to approximate functions and data distributions. 
In the literature we find the Gaussian and neural network combination proposed in many domains.
There are articles that use Gaussians as part of the neural network itself, such as~\cite{watanabe1996fuzzy}, where the Gaussians are used as activation functions, and~\cite{viroli2019deep}, where the layers of the neural network follow a Gaussian mixture model (GMM).
In general, the notable traits of a multivariate Gausssian is that knowing its parameters (means vector and covariance matrix) allow easy generation of new samples.
For example, in variational autoencoders~\cite{doersch2016tutorial} neural networks are used to estimate the parameters of a multivariate Gaussian; then the parameters are used to produce samples from the distribution of the data in the input space.
Another issue related to the combination of multivariate Gaussians and neural networks is how to integrate GMMs into neural networks in order to perform classification.  
In a recent study performed by Tüske \text{et al.}~\cite{tuske2015integrating} the authors took an approach similar to ours, proposing the integration of latent variables into the last layer of the neural network.
In their work, the last layer represents the parameters of the GMM. 
In this case, a GMM for the sample is calculated, and the prediction is made accordingly.
In our method, we 
approximate each class representation using a multivariate Gaussian distribution (a special type of GMM) on each class, but in contrast to~\cite{tuske2015integrating}, we do not incorporate the parameters of the Gaussians in the neural network, meaning that the parameters do not need to be learned explicitly, thus making the training easier.
One more important difference is that in the work performed by Tüske et al. ~\cite{tuske2015integrating} the authors use the same covariance matrix for all of the classes, while we approximate a different covariance matrix for each class.
\subsection{Large Margin Algorithms in Neural Networks}  
As shown by Boser \textit{at el.} ~\cite{boser1992training}, one of the desired qualities of a classifier is to have large margins between the representations of the classes. 
Liang \textit{at el.}~\cite{liang2017soft} explored the effects of large margins between the classes in neural networks.
Cross-entropy loss is currently the most frequent loss function used by neural network-based classifiers.
Sun \textit{et al.}~\cite{sun2016depth} empirically showed that the cross-entropy loss does not encourage a large margin. Neural networks contain many representations of the data, and hence, a reasonable approach for achieving the large margin effect in neural networks is to  form large margins in some of the network's inner representations.~\cite{liu2016large, sokolic2017robust, liang2017soft}.
In~\cite{elsayed2018large}, Elsayed et al. proposed a new optimization target aimed at replacing the softmax/cross-entropy combination completely. In our method, we also replace the softmax/cross-entropy combination - with a new optimization target - but our method forces a more general margin definition. The algorithms proposed by Elsayed et al. aim to maximize the distance between the closest points of different classes, while our method tries to maximize the distance between the means of each class, which represents a relaxation of the formal definition of a large margin between the classes.

\section{Methodology}
\label{section:three}
\begin{figure}[t]
\centerline{
\includegraphics[scale=0.47]{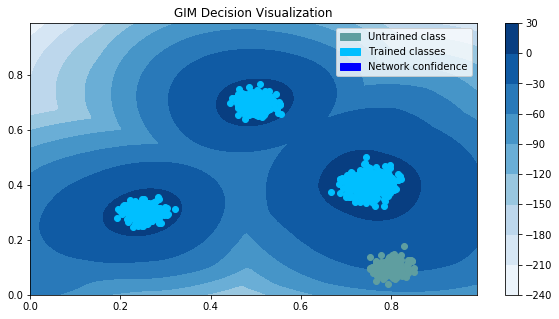}}
\caption{Gaussian isolation machine trained on three classes predicting the fourth class.}
\label{figure2}
\end{figure}

In this paper, we propose a means of improving the design of neural network classifiers so that they can cope with input that does not belong to one of the classes in the training set. The proposed classifier takes advantage of the fact that neural networks can model complex probability distributions to transform the input of the neural network into a vector space in which each class has an approximately known probability distribution - a Gaussian. 
More specifically, we train the neural network to produce an output space where the model's output has a simple distribution, thus forcing the samples from each class to behave like dense, isolated clusters in the output space. 
Forcing each class to be dense improves its approximation using a multivariate Gaussian with diagonal covariance matrix, and separating the clusters from one another is a way of creating large distances between the classes, which is similar to creating large margins.
In contrast to softmax-based neural networks that take a discriminative approach and directly model  $p(y|x)$, the Gaussian isolation machine models $p(y|x)$ as follows:

\begin{equation}
p(y|x) \approx p(y|f(x)) = \frac{ p(f(x)|y)\cdot p(y)}{p(f(x))}
\label{eq1}
\end{equation}

where $f(x)$ is the network output, and $y$ is the class label.
We use the likelihood probability component as our confidence rate, and this allows us to deal with data from untrained distributions.
The modeling technique used is similar to that of a generative model, but we consider it a hybrid model, because it is a discriminative model that is generative towards the output of the neural network and not toward the input space.
We formulated two ways of controlling the distribution of each class representation: the first one controls the class representation density (CTV loss), and the second one controls the class representation spread (CH loss).

\par
Consider a classification problem with $|C|$ classes, such that
$C= \{c_{1},\ldots,c_{|C|}\} $, where $c_{i}$ is the $i^{th}$ class, and a function $f:\chi \longrightarrow  \mathbb{R}^{d}$, such that $d$ is not necessarily equal to $|C|$, which represent a neural network.
We define the following metrics:

\begin{enumerate}
\item Class Mean Vector:
\begin{equation}
    \mu_{c} = \frac{1}{|c|}\sum_{x\in c}f(x) 
    \label{eq2}
\end{equation}
The mean vector of class $c\in C$ in the neural network output space.

\item Class Neighborhood Probability:
\begin{equation}
CNP(c_{1},c_{2}) = \exp{\frac{-\Vert \mu_{c_{1}} -\mu_{c_{2}} \Vert_{2}^{2}} {2\sigma_{c_{1}}^{2} }} 
\label{eq3}
\end{equation}
where $c_{1}$ and $c_{2}$ are classes, and $\mu_{c_{i}}$ is the mean vector of each class (Equation \eqref{eq2}).
This metric corresponds to the unnormalized probability of class $c_{1}$ being near class $c_{2}$, assuming a Gaussian distribution on class $c_{1}$ with a diagonal covariance matrix whose diagonal elements are all equal to $\sigma_{c_{1}}^{2}$. 
For optimization purposes, the class neighborhood probability equation (Equation \eqref{eq3}) is insufficient for the purpose of separating the class and achieving the large margin effect, because when the probability is low, there isn't a need for the network to separate the classes. Therefore, for optimization, we use the following modified version of the equation to ensure class separation and the desired large margin effect:
 \begin{equation}
\Theta(c_{1},c_{2}) = \exp{\frac{-\Vert \mu_{c_{1}} -\mu_{c_{2}} \Vert_{2}^{2}} {2\cdot\alpha\cdot\sigma_{c_{1}}^{2} }} \label{eq4}
\end{equation}
where $\alpha$ is a large constant, compared to the actual class covariance matrix diagonal elements. This is similar to the method used by Hinton, et al   \cite{hinton2015distilling}, but it is used for a  different purpose. In our method, during optimization this constant forces the assumed Gaussians to (1) cover more space, and (2) always include the other classes.

\item Center Distance: 
\begin{equation}
D(x,c) = \Vert \mu_{c}-f(x) \Vert_{2}^{2}
\label{eq5}
\end{equation}
where $\mu_{c}$ is the class mean vector (Equation(\eqref{eq2}). This metric is the $l_{2}$ distance from the class mean.

\item Center loss~\cite{wen2016discriminative} as the Class Total Variance:\\
The class total variance (CTV) is the first moment of the center distance (Equation \eqref{eq5}) over the samples of class $c \in C$:
\begin{equation}
CTV(c) = \frac{1}{|c|}\sum_{x\in c}D(x,c) =  \frac{1}{|c|} \sum_{x\in c} \Vert \mu_{c}-f(x) \Vert_{2}^{2}
 \label{eq6}
\end{equation}

The class total variance is equal to the sum of the diagonal elements of the covariance matrix of class $c \in C$:
%  $$
%  =\frac{1}{|c|} \sum_{i\in c} (x^{T}_{i}x_{i} -2x^{T}_{i}\mu_{c} +\mu_{c}^{T}\mu_{c}) =
%  $$
\newline
\newline
$CTV(c)=$ \\
 $$
=\frac{1}{|c|} \sum_{x\in c} f(x)^{T}f(x) 
 - \frac{2}{|c|} \sum_{x\in c} f(x)^{T}\mu_{c}
 + \frac{1}{|c|} \sum_{x\in c} \mu_{c}^{T}\mu_{c} 
 $$
%  $$
%  \frac{1}{|c|} \sum_{i\in c} x^{T}_{i}x_{i} 
%  - \frac{2}{|c|} \sum_{i\in c} \sum_{j \leq d} x_{i,j}\mu_{c}_{j}
%  + \frac{1}{|c|} \sum_{i\in c} \mu_{c}^{T}\mu_{c}
%  $$
 $$
  =\frac{1}{|c|} \sum_{x\in c} f(x)^{T}f(x) 
 - \sum_{j\leq d} \mu_{c,j} \frac{2}{|c|} \sum_{x \in c} f(x)_{j}
 +\mu_{c}^{T}\mu_{c} 
 $$
%  $$
%   \frac{1}{|c|} \sum_{i\in c} x^{T}_{i}x_{i} 
%  - 2\mu_{c}^T\mu_{c} + \mu_{c}^{T}\mu_{c} =
%  $$
 $$
 =\frac{1}{|c|} \sum_{x\in c} f(x)^{T}f(x) 
- \mu_{c}^T\mu_{c} 
$$
$$
=\sum_{j\leq d} Diag( \frac{1}{|c|} \sum_{x \in c} f(x)f(x)^{T}-\mu_{c}\mu_{c}^T)_{j}
 $$

Class Homogeneity:\\
\begin{equation}
    CH(c) = \frac{1}{|c|} \sum_{x\in c}(CTV(c)- D(x,c))^{2}
    \label{eq7}
\end{equation}
The class homogeneity (CH) is the second moment of the Equation \eqref{eq5} over the samples of class~$c \in C$, and it defines the variance of distances from the center of the class.

\end{enumerate}

The minimization of Equation \eqref{eq6} will effectively shrink the diagonal elements of the covariance matrix of the class in the output space, thus making its representation small and dense. This minimization allows us to assume a multivariate Gaussian distribution for each class with a diagonal covariance matrix, those diagonal covariance matrices will be employed when making predictions.

In the first of the two optional loss functions for the GIM, the combination of Equations \eqref{eq6} and \eqref{eq4} results in the CTV loss. This loss function both controls the class representation size in the output space and ensures that the representations of classes  will be far apart from one another.\\ 
 \begin{equation}
     \frac{\lambda}{|C|}\sum_{c\in C} CTV(c)+
     \frac{1}{|C|^{2}}\sum_{c\in C}\sum_{l\in C}\Theta(c,l)
     \label{eq8}
 \end{equation}

When combining the class neighborhood probability  (Equation \eqref{eq4}) and the class homogeneity (Equation \eqref{eq7}), we were able to attain the second loss function for the GIM - the CH loss. This loss function controls the class spread by ensuring that the variance of $l_{2}$ distances from the class mean vector will be small \\
\begin{equation}
 \frac{\lambda}{|C|}\sum_{c\in C} CH(c)+
 \frac{1}{|C|^{2}}\sum_{c\in C}\sum_{l\in C}\Theta(c,l)
 \label{eq9}
\end{equation}
We trained individual neural networks to minimize Equations \eqref{eq7} and \eqref{eq9}), using different neural network architectures where the last layer is an arbitrary size (e.g., 24, 32, 64). As we hoped, the distribution of the network's output is similar to that of a GMM, with a large Euclidean distance between the clusters (classes).

Knowing that the output space approximately distributed as a multivariate Gaussian provides us with access to the likelihood term $p(f(x)|c)$, which can be thought of as a confidence metric for the neural network's predictions. The likelihood probability is the probability for a sample $x$ to belong to a certain class $c\in C$ in the neural network's output space, and it is defined as follows: (the multivariate Gaussian probability density function):\\
 \begin{equation}
     p(f(x)|c) =  (2\pi)^{-0.5d}\cdot |\Sigma_{c}|^{-0.5}
     \cdot\exp\{-\frac{1}{2}\cdot\Vert \mu_{c} - f(x) \Vert_{\Sigma^{-1}_{c}}^{2}\}
     \label{eq10}
 \end{equation}
\par
In practice, we use the $\log()$ of the probability.
We use the log-likelihood as a confidence metric that allows us to differentiate between in-distribution data and out-of-distribution data, by setting a threshold on its value, so that inputs that result in a value lower then this confidence threshold  will be labeled as out-of-distribution.
\par
To make a prediction, the GIM follows an approach similar to that of many generative classifiers. It uses a term which includes both the likelihood term and the prior over the labels.
We combine the confidence (log-likelihood) term with a prior over the class labels and utilize Bayes' rule to approximate $p(y|x)$ as follows:\\
Prior over the labels:\\
\begin{equation}
    p(c) = \frac{|c|}{\sum_{c'\in C} |c'|}
    \label{eq11}
\end{equation}
Posterior probability for classification:\\
\begin{equation}
\begin{split}
\underset{c}{argmax}~p(c|x) \approx \underset{c}{argmax}~  p(c|f(x)) 
= & \\  \underset{c}{argmax}~\frac{p(f(x)|c)\cdot p(c)}{p(x)} \propto\\ \underset{c}{argmax}~p(f(x)|c)\cdot p(c)
\end{split}
\label{eq12}
\end{equation}
\par
In practice, we use the $log()$ of the whole expression, in order to avoid numerical errors.
As can be seen in Figure~\ref{figure2}, our method has a different decision boundary shape, so that rather than splitting the input space into three areas (as seen in Figure~\ref{figure1}), a heat map is created for each class probability distribution.

\section{Evaluation}
\label{section:four}
\subsection{Experimental Settings}
\begin{table}[b]
\caption{Neural Network Architectures}
\begin{center}
\resizebox{0.9\textwidth}{!}{\begin{minipage}{\textwidth}

\begin{tabular}{|c|c|c|c|}
\hline
\textbf{Layer}&\multicolumn{3}{|c|}{\textbf{Neural Network Architectures}} \\
\cline{2-4} 
\textbf{Number} & \textbf{\textit{MNIST}}& \textbf{\textit{FASHION-MNIST}}&
\textbf{\textit{IMDB}} \\

\hline
1& Conv2D 3X3X32& Conv2D 3X3X32& Embedding \\
\hline

\hline
2& Conv2D 3X3X64& Conv2D 3X3X32& Dropout 0.25 \\
\hline

\hline
3& MaxPool 2X2& MaxPool 2X2,strides 2& Conv1D 250X3 \\
\hline

\hline
4& Dropout 0.25& Dropout 0.3&  GlobalMaxPool1D\\
\hline

\hline
5& Flatten & Conv2D 3X3X64&  Dense 250\\
\hline

\hline
6& Dense 32& Conv2D 3X3X64&  \\
\hline

\hline
7& & MaxPool 2X2, strides 2&  \\
\hline

\hline
8& & Dropout 0.4&  \\
\hline

\hline
9& & Flatten &  \\
\hline

\hline
10& & Dense 24 &  \\
\hline

\end{tabular}%
\label{table1}
\end{minipage}}
\end{center}
\end{table}
In this section, we evaluate the performance of the GIM on three tasks and compare its performance to standard neural networks. 
Our evaluation shows that the Gaussian isolation machine achieves similar classification results and has a similar convergence speed to that of standard neural networks, while possessing the inherent ability to detect OOD data with high accuracy.
We evaluate the GIM on two classic classification tasks and three out-of-distribution data detection tasks.
For classification, we chose three standard object detection benchmark datasets and one sentiment analysis benchmark dataset. 
In all of the classification experiments we compare our method to state-of-the-art neural network classifiers with the same architecture, but we removed the last layer (weights and softmax activation) of the GIMs, and as a result, they have fewer parameters. 
We created two scenarios for the identification of untrained distribution data (OOD) experiments. 
In these experiments, we trained a GIM on several classes of a dataset and determined whether data from the remaining classes is classified by the GIM as one of the trained classes. 
In addition, we also performed an experiment similar to the one presented in Linag \textit{et al.}~\cite{liang2017enhancing} to compare the GIM's detection abilities to the baseline detector~\cite{hendrycks2016baseline}.
To measure the classification accuracy and convergence speed, we used the architectures presented in Table~\ref{table1}.
For the CIFAR 10 dataset, we trained a ResNet20v1, like the one presented in~\cite{he2016deep}, which is a very compact ResNet with under $300K$ trainable parameters. We used Keras~\cite{chollet2015keras} to implement the neural network, but because our implementation of ResNet20V1 uses random data augmentation, we don't obtain the same results as those presented in ~\cite{he2016deep}. 
However, after several trials we were able to achieve 91.2\% accuracy, matching the results in that paper. 
In addition to the ResNet20v1, we  also trained a VGG16~\cite{simonyan2014very}, without its final layer.

\subsection{Classification Accuracy}
\begin{table}[b]
\caption{Gaussian Isolation Machine vs Standard Neural Networks}
\begin{center}
\resizebox{0.95\textwidth}{!}{\begin{minipage}{\textwidth}

\begin{tabular}{|c|c|c|c|}
\hline
\textbf{Dataset}&\multicolumn{3}{|c|}
{\textbf{GIM vs Standard }} \\
\cline{2-4} 
& \textbf{\textit{GIM-CTV equation}}& \textbf{\textit{GIM-CH equation}}&
\textbf{\textit{Standard}} \\

\hline
MNIST& 98.6 & \textbf{99.25} & 99.08 \\
\hline

\hline
FASHION& 92.22& 93.05 &\textbf{ 93.55} \\
\hline

\hline
IMDB& 89.0 & 89.4& \textbf{89.5}\\
\hline

\hline
CIFAR 10-ResNet& $89.5$& 89.4& \textbf{$89.7$}\\
\hline

\hline
CIFAR 10-VGG& \textbf{93.5} & 93.4 & \textbf{93.5}\\
\hline

\end{tabular}
\label{table2}
\end{minipage}}
\end{center}
\end{table}

In this section, we determine the accuracy of the GIM and compare it to that of standard neural networks.
All of the neural networks were created using the architectures presented in Table~\ref{table1} and were initialized using the same random seed. We test our method on the MNIST character recognition dataset ~\cite{lecun-mnisthandwrittendigit-2010}, the FASHION-MNIST clothing recognition dataset ~\cite{xiao2017fashion}, the CIFAR 10 object detection dataset  ~\cite{krizhevsky2009learning}, and the IMDB sentiment analysis dataset ~\cite{maas-EtAl:2011:ACL-HLT2011}.
In Table \ref{table2}, we provide a comparison of the results, presenting the average accuracy achieved by each method with each dataset. 
It is clear from the results presented in Table~\ref{table2} that our method does not compromise the classification accuracy and that in some cases it even improves it.
When examining other generative and hybrid models, such as Bayesian neural networks~\cite{gal2015bayesian}, KNN, Na\"ive Bayse, and ClassRBM~\cite{larochelle2012learning}, the accuracy level is usually low when the datasets contain high dimensional data, as is the case with the MNIST and CIFAR10 images.
The novelty of our work is that we were able to retain the classification accuracy of the fully discriminative neural network, while creating a hybrid model.  
\subsection{Convergence Speed}
 \begin{figure}[t]
\centerline{
\includegraphics[scale=0.47]{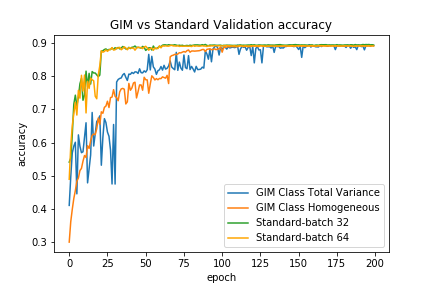}}
\caption{The Gaussian isolation machine vs a standard neural network on the CIFAR 10 dataset.}
\label{figure3}
\end{figure}
Figure~\ref{figure3} compares the convergence speed of the GIM to that of a standard neural network, where both methods use the ResNet20v1 architecture. All training was performed using a single NVIDIA-2080 TI GPU. Note that both formulations of the Gaussian isolation machine converge slower than a standard neural network, although they achieved nearly the same final accuracy (see Table \ref{table2}). We hypothesize that the slower convergence speed is due to the fact that the GIM must separate the representations of the classes, as well as isolate them and force them into a Gaussian form.

\subsection{Identifying Out Of Distribution Data}
\begin{figure}[b]
\centerline{
\includegraphics[scale=0.47]{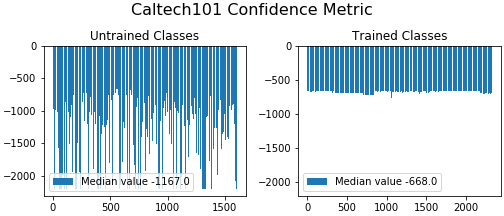}}
\caption{Caltech101 dataset confidence metric for the trained and untrained class data.}
\label{figure4}
\end{figure}

\begin{figure}[ht]
\centerline{
\includegraphics[scale=0.47]{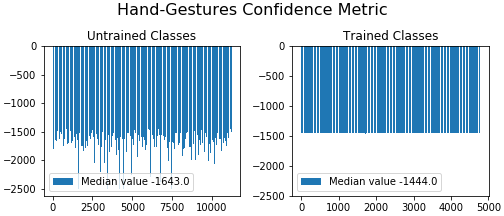}}
\caption{Hand-Gesture dataset  confidence metric for the trained and untrained class data.}
\label{figure5}
\end{figure}

Anomalous data and data from outside the trained distributions can appear in a variety of applications. The proposed method can detect data from other distributions, i.e., data classes that the model did not train on. In this section, we empirically evaluate the proposed method's ability to distinguish between data from the trained distribution and data from outside the trained distribution.
To accomplish this, we designed two experiments in which we trained a GIM on a portion of the classes in a dataset and evaluated its detection ability on the remainder of the classes in the dataset. 
In the first experiment, we used the Caltech101 dataset~\cite{fei2004learning}, and in the second experiment, we used the Hand-Gestures~\cite{mantecon2016hand} dataset. In both experiments, we also trained standard neural networks with the same architecture and compared our performance to those of the neural networks.
\par
The difference between the confidence values for the GIM's predictions for Caltech101 data from the trained distribution (first 10 classes) and Caltech101 data from untrained distribution (classes 10-40) can be seen in Figure~\ref{figure4}.  There is a big difference between the two graphs presented in the figure: data from classes that our model trained on has much higher confidence than data from classes it didn't train on. This observation makes it possible to set a threshold for the confidence values and, given an input, determine whether it belongs to the trained distribution or not. In this case, the threshold value was set at $-665$, creating an almost perfect separation of $99.8\%$ between out-of-distribution data and the trained distribution. In the second experiment, we loaded the Hand-Gestures dataset and resized it to $128\times128$. The GIM trained on the first three classes. The confidence values can be seen in Figure~\ref{figure5}. Here we used a confidence threshold of $-1450$. Again, the GIM achieved an almost perfect detection rate of $ \approx97.5\%$. 
\par
To perform a fair comparison to other OOD detectors, we implemented the baseline method introduced by Hendrycks \& Gimpel\cite{hendrycks2016baseline}, comparing the baseline method to the GIM when the threshold values for the softmax (baseline) and the log-likelihood (GIM) were set such that the $TPR(TP/(TP+FN))$ would yield $97\%$, i.e., 97\% of the neural networks' predictions on the training set will be above the thresholds.

Table~\ref{table6} presents an evaluation similar to that presented by Liang et al. ~\cite{liang2017enhancing}.
A comparison is made between two VGG13 neural networks trained on the CIFAR 10 dataset, to $\approx93\%$ test accuracy. The CIFAR 10 test set serves as the in-distribution data, and the out-of-distribution data comes from the following datasets: Tiny ImageNet, LSUN, iSUN.
The results of this comparison appears in Table~\ref{table6}.
\begin{table}[t]
\caption{Baseline Method vs Gaussian Isolation Machine Detection of OOD Data (the CIFAR 10 test set serves as the in-distribution data)}
\resizebox{0.95\textwidth}{!}{\begin{minipage}{\textwidth}

\begin{tabular}{|p{.06\textwidth}|c|c|c|c|c|}
\hline
\multicolumn{6}{|c|}{\textbf{Baseline (Hendrycks \& Gimpel, 2017) / GIM}} \\ \hline
\textbf{OOD Dataset} & \textbf{FPR} & \textbf{\begin{tabular}[c]{@{}c@{}}Detection\\ Error\end{tabular}} & \textbf{AUROC} & \textbf{\begin{tabular}[c]{@{}c@{}}AUPR\\ In\end{tabular}} & \textbf{\begin{tabular}[c]{@{}c@{}}AUPR\\ Out\end{tabular}} \\ \hline
\textbf{ImageNet (resize)} &0.70/\textbf{0.21}  &  0.36/\textbf{0.16}  & 0.85/\textbf{0.86}  &   \textbf{0.88}/0.80     & 0.82/\textbf{0.86}   \\ \hline
\textbf{ImageNet (crop)}   & 0.57/\textbf{0.13} &  0.33/\textbf{0.12}  & 0.90/\textbf{0.92}  &   \textbf{0.92/0.92}     & 0.88/\textbf{0.89}   \\ \hline
\textbf{LSUN (resize)}     & 0.54/\textbf{0.17 }&  0.29/\textbf{0.14 } & 0.81/\textbf{0.87}  &    \textbf{0.85}/0.79    &  0.76/\textbf{0.87}  \\ \hline
\textbf{LSUN (crop)}       & 0.68/\textbf{0.18} &  0.34/\textbf{0.15}  & 0.89/\textbf{0.91}  &  \textbf{ 0.92/0.92}     &  0.85/\textbf{0.88}  \\ \hline
\textbf{iSUN}              & 0.80/\textbf{0.15} &  0.42/\textbf{0.14}  & 0.76/\textbf{0.82}  &  \textbf{0.84}/0.79      &  0.74/\textbf{0.87}  \\ \hline
\textbf{Gaussian}          &  1.0/ \textbf{0.0} &  0.52 /\textbf{0.05} & 0.84/\textbf{0.99}  &  0.97/\textbf{0.99}      &  0.83/\textbf{0.94}  \\ \hline
\textbf{Uniform}           & \textbf{0.0/0.0}   &   \textbf{0.02}/0.05 & 0.91/\textbf{0.99 } &   0.97/\textbf{0.99}     &  0.83/\textbf{0.94}    \\ \hline
\end{tabular}
\label{table6}
\end{minipage}}
\end{table}

\subsubsection{Evaluation Metrics}
\begin{itemize}
\item \textbf{TPR and FPR}: Measures the false positive rate and true positive rate. Let TP,  FP,  TN, and FN respectfully represent the true positives, false positives, true negatives, and false negatives. The true positive rate is calculated as $TPR = TP/(FP+TP)$, and the false positive rate is calculated as $FPR = FP/(FP+TN)$.
\item \textbf{AUROC}:  Measures the area under the ROC curve. The receiver operating characteristic (ROC) curve plots the relationship between the TPR and
          FPR. The area under the ROC curve can be interpreted as
          the probability that a positive example (in-distribution) will
          have a higher detection score than a negative example (out-of-distribution).
\item \textbf{AUPR}: Measures the area under the precision-recall (PR)
          curve. The PR curve is created by plotting precision =
          TP/(TP + FP) against recall = TP/(TP + FN). In our
          tests, AUPR In denotes in-distribution data, which is used as the positive class, and AUPR Out denotes out-of-distribution examples, which are used as the positive class.

\end{itemize}
\subsubsection{Test Sets for OOD Detection}
\begin{itemize}
    \item \textbf{Tiny ImageNet}
    This is a subset of the original ImageNet dataset containing 200 classes. For testing purposes we used two datasets that were created from the Tiny ImageNet test set which contains 10,000 images: ImageNet (resize) and ImageNet (crop).
    \item \textbf{LSUN}
    The Large-scale Scene Understanding dataset contains 10,000 test images which were used to create two datasets: LSUN (resize) and LSUN (crop).
    \item \textbf{iSun}
    The iSUN dataset is a subset of the
    SUN dataset, and it contains 8,925 images. All images in this
    dataset were used, resized to 32 × 32 pixels.
    \item \textbf{Gaussian and Uniform Noise}
    The Gaussian and Uniform Noise datasets are datasets created by sampling 10,000 $32\times 32$ pixel images from a uniform distribution and 10,000 $32\times 32$ pixel images sampled $i.i.d.$ from 2D multivariate Gaussian distribution with a mean of 0.5 and STD of one. 
\end{itemize}

% \subsection{Regularization Case Study-CVC Regularization}
% \input{Regularization.tex}

\section{Summary}
\label{section:five}
In this paper, we presented the Gaussian isolation machine, a new neural network-based classification method. The GIM is based on neural network that was trained to transfer the inputs to a vector space where the data distribution can be approximated using multivariate Gaussian.
The approach integrates principles from generative and discriminative models to form a hybrid classification method that can classify data with high accuracy, as well as to identify data from untrained distributions. In the process of creating the Guassian isolation machine, we also experimented with new regularization terms that improves the classification ability of cross-entropy/softmax-based neural networks. The main contribution of this paper is the ability of the GIM to identify whether the input data is from the training set distribution or not, without the need for any preprocessing or external detection measures. 
In future work, we intend to add a sampling capability to the GIM (i.e.,  giving it the ability to produce new samples), and make changes to the loss function to enable it to perform multi-label classification.
In our experiments we tried using the full covariance matrix of each class, and found that although the classification results were better, the run-time was much longer. We believe that additional research in this area will lead to better classification results.

\bibliography{conference.bib}
\bibliographystyle{IEEEtran}
\end{document}